%% file: naacl.tex
\documentclass[11pt,a4paper,draft]{article}
\usepackage{acl2020, times, latexsym}

\input{preambles/base.tex}

\input{preambles/mathops.tex}

\input{preambles/theorems.tex}

\input{preambles/tikz.tex}

\usepackage[varg]{txfonts}

\usepackage{booktabs, multirow, siunitx}
\definecolor{lightlightgray}{gray}{0.9}

\usepackage{ellipsis}
\newcommand\narrowdots{\mbox{\renewcommand{\ellipsisgap}{0.01em}\dots}}
\newcommand\cnarrowdots{{\renewcommand{\ellipsisgap}{0.01em}\mathinner{{\cdotp}\kern\ellipsisgap{\cdotp}\kern\ellipsisgap{\cdotp}}}}

\usepackage{standalone, subcaption, colortbl}

\aclfinalcopy 

\title{Supertagging-based Parsing\\with Linear Context-free Rewriting Systems}

\author{%
  Richard Mörbitz \and Thomas Ruprecht \\
  Faculty of Computer Science \\
  Technische Universität Dresden \\
  01062 Dresden, Germany \\
  {\tt $\{$richard.moerbitz$,$thomas.ruprecht$\}$@tu-dresden.de}%
}

\date{}

\begin{document}
\maketitle
\begin{abstract}
    We present the first supertagging-based parser for LCFRS.
    It utilizes neural classifiers and tremendously outperforms previous LCFRS-based parsers in both accuracy and parsing speed.
    Moreover, our results keep up with the best (general) discontinuous parsers, particularly the scores for discontinuous constitutents are excellent.
    The heart of our approach is an efficient lexicalization procedure which induces a lexical LCFRS from any discontinuous treebank.
    It is an adaptation of previous work by \citet{MoeRup20}.
    We also describe a modification to usual chart-based LCFRS parsing that accounts for supertagging and introduce a procedure for the transformation of lexical LCFRS derivations into equivalent parse trees of the original treebank.
    Our approach is implemented and evaluated on the English Discontinuous Penn Treebank and the German corpora NeGra and Tiger.
\end{abstract}

\input{figures/discontinuous.tex}

\section{Introduction}

Constituency parsing is a syntactical analysis in NLP that aims to enhance sentences with, usually tree-shaped, phrase structures (for an example cf.\@ the left of Fig.~\ref{fig:discontinuous}).
Formalisms such as context-free grammars (CFG) are used in this setting because they are conceptually simple, interpretable, and parsing is tractable (cubic in sentence length).

Discontinuous constituents span non-contiguous sets of positions in a sentence.
The resulting phrase structures do not take the shape of a tree anymore, as they contain crossing branches (cf.\@ the left of Fig.~\ref{fig:discontinuous}), and cannot be modeled by CFG.
As a countermeasure, many corpora (e.g.\@ the Penn Treebank \citep[PTB]{Marcus94}) denote these phrase structures as trees nevertheless and introduce designated notations for discontinuity, which is then often ignored in parsing.
However, discontinuity occurs in about 20\,\% of the sentences in the PTB and to an even larger extent in German treebanks such as NeGra and Tiger.
For parsing discontinuous constituents, so-called \enquote{mildly context-sensitive} grammar formalisms have been investigated, e.g.\@ tree-adjoining grammars (TAG; \citealp{JosLevTak75}) and linear context-free rewriting systems (LCFRS; \citealp{VijWeiJos87}).
These approaches have been empirically shown to indeed improve accuracy (cf.\@ e.g.\@ \citealp{EvaKal11}).
However, their increased expressiveness comes at the cost of a higher parsing complexity:
given a sentence of length $n$, parsing is in $O(n^6)$ for TAG and $O(n^{3\cdot\operatorname{fanout}(G)})$ for an LCFRS $G$.
The fanout is grammar-specific and reflects the degree of discontinuity in the rules of $G$.
The expressiveness of TAG equals that of LCFRS with fanout 2.
An LCFRS derivation of a discontinuous phrase is shown in the right of Fig.~\ref{fig:discontinuous}.

One approach for making parsing with mildly context-sensitive grammars tractable is \emph{supertagging}, which was originally introduced for lexical TAG \cite{bangalore1999supertagging}.
A TAG is lexical if each rule contains one word.
The supertagger is a (often discriminative) classifier that selects for each position of the input sentence a subset of the rules of the TAG; these are the so-called supertags.
Parsing is then performed with the much smaller grammar of supertags.
Research on supertagging has also been conducted in the context of combined categorical grammars \cite{clark2002supertagging}, but not yet for LCFRS.
Recently, the use of recurrent neural networks (RNN) as classifiers in supertagging has improved not only the speed of such systems, but also their accuracy by far \cite{vaswani2016supertagging,Kas17,Bla18,Kad18}.

In this paper, we present the first supertagging-based parser for LCFRS.
We adapt the lexicalization procedure of \citet{MoeRup20} to efficiently induce a lexical LCFRS from any given treebank.
We modify the usual chart-based parsing approach for LCFRS to account for supertagging and introduce a procedure which transforms lexical LCFRS derivations into equivalent parse trees of the original treebank.
We implemented the approach and show that it vastly outperforms other LCFRS-based parsers in both accuracy and parsing time on three discontinuous treebanks (one English and two German).
Our results can keep up with recent state-of-the-art discontinuous constituent parsers, most notably the parsing scores of discontinuous constituents are excellent.
The implementation of our approach will be published on GitHub.

\section{Preliminaries}
The set of \emph{non-negative} (resp.\@ \emph{positive}) \emph{integers} is denoted by \(\DN\) (resp.\@ \(\DN_+\)).
We abbreviate \(\{1, \narrowdots, n\}\) by \([n]\) for each \(n \in \DN\).
Let \(A\) be a set; the \emph{set of (finite) strings over \(A\)} is denoted by \(A^*\).
An \emph{alphabet} is a finite and non-empty set.

Let \(S\) be some set whose elements we call \emph{sorts}.
An \emph{\(S\)-sorted set} is a tuple \((A, \mathit{sort})\) where \(A\) is a set and \(\mathit{sort}: A \to S\).
Usually, we identify \((A, \mathit{sort})\) with \(A\). We denote \(\mathit{sort}\) by \(\sort_A\) and the set \(\mathit{sort}^{-1}(s)\) by \(A_s\) for each \(s \in S\).
We use set notation (\(\in, \subseteq, \cup, \ldots\)) with sorted sets in the intuitive manner.
Now let \(A\) be an \((S^* \times S)\)-sorted set.
The \emph{set of trees over \(A\)} is the \(S\)-sorted set \(\sTrees^A = \cup_{s \in S} \sTrees^A_s\) where \(
  \sTrees^A_s = \{a(t_1, \ldots, t_k) \mid k \in \DN, s_1, \ldots, s_k \in S, a \in A_{(s_1 \cdots s_k, s)}, t_1 \in \sTrees^A_{s_1}, \ldots, t_k \in \sTrees^A_{s_k}\}
\) for each \(s \in S\).
A \emph{ranked set} \(A\) is an \((S^* \times S)\)-sorted set where \(S = \{s\}\); the notation \(\rank_A(a) = k\) abbreviates \(\sort_A(a) = (s^k, s)\), and \(A_k\) abbreviates \(A_{(s^k, s)}\).
If we use a usual set \(B\) in place of a ranked set, we will silently assume \(\rank_B(b) = 0\) for each \(b \in B\).
Let $X$ be a set.
We let $A(X) =  \{ a(x_1, \dots, x_k) \mid k \in \mathbb N, a \in A_k, x_1, \dots, x_k \in X \}$.

\paragraph{LCFRS.}
\emph{Linear context-free rewriting systems} extend the rule-based string rewriting mechanism of CFG to string tuples; we describe the generation process by \emph{compositions}.
Let \(k \in \DN\) and \(s_1, \ldots, s_k, s \in \DN_+\); a \emph{\(\Sigma\)-composition} is a tuple \((u_1, \ldots, u_s)\) where each \(u_1, \ldots, u_s\) is a non-empty string over \(\Sigma\) and variables of the form \(\x_i^j\) where \(i \in [k]\) and \(j \in [s_i]\).
Each of these variables must occur exactly once in \(u_1 \cdots u_s\) and they are ordered such that \(\x^1_{i}\) occurs before \(\x^1_{i+1}\) and \(\x^j_i\) occurs before \(\x^{j+1}_i\) for each \(i \in [k-1]\) and \(j \in [s_i-1]\).
We denote the set of $\Sigma$-compositions by \(\C^\Sigma_{(s_1 \cdots s_k, s)}\); we drop the superscript in the case \(\Sigma = \emptyset\) (then $C_{(s_1 \cdots s_k, s)}$ is finite); we drop the subscript if we admit any configuration of \(k\), \(s_1, \ldots, s_k\) and \(s\).
We associate with each composition \((u_1, \ldots, u_s) \in \C^\Sigma_{(s_1\cdots s_k,s)}\) a function from \(k\) string tuples, where the \(i\)-th tuple is of length \(s_i\), to a string tuple of length \(s\).
This function is denoted by \(\sem{(u_1, \ldots, u_s)}\).
Intuitively, it replaces each variable of the form \(\x_i^j\) in \(u_1, \ldots, u_s\) by the \(j\)-th component of the \(i\)-th argument.

Let \(c \in \C^\Sigma_{s_1 \cdots s_k, s} \) be a composition and \(\sigma \in \Sigma\).
If \(s_i = 1\) for some \(i \in [k]\), we obtain the \emph{$i$-partial application of \(c\) to \(\sigma\)}, denoted by \(\sem{c}_i(\sigma) \in \C^{\Sigma}_{s_1 \cdots s_{i-1} s_{i+1} \cdots s_k, s}\), from \(c_2\) by replacing \(\x_i^1\) by \(\sigma\), and each \(\x_{\hat{i}}^j\) by \(\x_{\hat{i}-1}^j\) for \(\hat{i} > i\).

For each \(s \in \mathbb{N}\), we denote the composition \((\x_1, \ldots, \x_s)\) by \(\id_s\).

An \emph{LCFRS} is a tuple \(G=(N, \Sigma, S, R)\) where
\begin{itemize*}
  \item \(N\) is a finite \(\DN_+\)-sorted set (\emph{nonterminals}),
  \item \(\Sigma\) is an alphabet (\emph{terminals}),
  \item \(S \in N_1\) (\emph{initial nonterminal}), and
  \item \(R\) is a finite \((N^* \times N)\)-sorted set (\emph{rules}).
     Each rule is of the form \(A \to c(B_1, \ldots, B_k)\), where \(k \in \DN\), \(A, B_1, \ldots, B_k \in N\), and \(c \in \C^\Sigma_{(\sort_N(B_1) \cdots \sort_N(B_k), \sort_N(A))}\).
     The sort of the rule is \((B_1 \cdots B_k, A)\); we call \(A\) the \emph{left-hand side} (\emph{lhs}), \(B_1, \ldots, B_k\) the \emph{right-hand side} (\emph{rhs}) and \(c\) the rule's \emph{composition}.
     We drop the parentheses around the rhs if \(k = 0\).
\end{itemize*}
We call rules of the form
\begin{itemize*}
  \item \(A \to c\) where \(A \neq S\) \emph{terminating},
  \item \(A \to c(B)\) \emph{monic}, and
  \item \(A \to c(B_1, \narrowdots, B_k)\) where \(k\geq2\) \emph{branching}.
\end{itemize*}
A rule is called \emph{(uni-/double-)lexical}, if its composition contains \emph{at least one} terminal (resp.\@ \emph{exactly one} terminal/\emph{exactly two} terminals).
The LCFRS \(G\) is called \emph{(uni-/double-)lexical}, if each rule is (uni-/double-)lexical.
The \emph{set of (complete) derivations in \(G\)} is \(\derivs^G = \sTrees^R_S\).
Let \(d = r(d_1, \ldots, d_k) \in \sTrees^R\) with \(r = A \to c(B_1, \ldots, B_k)\).

\section{Obtaining Lexical LCFRS}\label{sec:lex}%

We extract an uni-lexical LCFRS from a discontinuous corpus using a lexicalization scheme similar as described by \citet{MoeRup20}.

Given a corpus, the procedure that induces an uni-lexical LCFRS from it is roughly as follows.
\begin{enumerate}
    \item Binarize each tree in the corpus.
        \label{step:short:binarize}
    \item Transform each tree to an LCFRS derivation using the standard technique for induction of LCFRS \cite{MaierSogaard08}.
        \label{step:short:readoff}
    \item Collapse every chain of monic rules; the nonterminals of each chain are combined to a new nonterminal.
        \label{step:short:dechain}
    \item Insert the lexical symbol of each non-initial terminating rule into its parent, then remove this rule.
        \label{step:short:fuseterm}
    \item Each branching rule has a distinct double-lexical terminating rule as a successor.
        Remove one terminal from the double-lexical rule and propagate it up the entire path to the branching rule.
        \label{step:short:propterm}
    \item Split all remaining double-lexical terminating rules into an uni-lexical monic rule and an uni-lexical terminating rule.
        \label{step:short:split}
    \item Read off the rules of each derivation.
        They are the rules of the uni-lexical LCFRS $\lexG$.
\end{enumerate}

Steps~\ref{step:short:dechain}--\ref{step:short:propterm} correspond to a variation of the lexicalization scheme by \citet{MoeRup20}.
The approach described here differs from the previous one in that it works on individual derivations rather than on an entire grammar.
As a consequence, lexical LCFRS obtained at the end may have a different language, but it is much smaller, which benefits the supertagging approach.

In the following, we describe the steps~\ref{step:short:dechain}--\ref{step:short:split} in very detail and show examples in figs.\@ \ref{fig:dechain}--\ref{fig:split}.
After step~\ref{step:short:readoff}, we obtain derivations of a \emph{binary}, \emph{terminal-} and \emph{initial-separated} LCFRS, i.e.\@ each occurring rule is either of the form \(A \to (\sigma)\), where $\sigma$ is a lexical symbol and \(A\) a part-of-speech tag, \(A \to c(B_1)\), or \(A \to c (B_1, B_2)\) where \(c\) contains no terminal symbols and none of \(B_1, B_2\) is the initial nonterminal.
Furthermore, we assume the composition of each monic rule \(A \to c (B)\) to be \(c = \id_{\fanout(B)}\).
Let $d$ be such a derivation.

\paragraph{Step~\ref{step:short:dechain}.}
We \emph{repeatedly} replace parts in \(d\) of the form
  \(A \to \id_{\fanout(B)} (B) \Big( B \to c \Big) \) by \( A\text+B \to c \), and
  \(A \to \id_{\fanout(B)} (B) \Big( B \to c(C_1, C_2) \Big( \ldots \Big) \Big) \) by \( A\text+B \to c (C_1, C_2) \Big( \ldots \Big) \),
until there is no monic rule in \(d\) left.
If the rule \(A \to \id_{\fanout(B)} (B)\) has a parent in \(d\), then the nonterminal \(A\) in its rhs is also replaced by \(A\text+B\).
After this step, there are only braching rules and terminating rules in \(d\).
Fig.\@ \ref{fig:dechain} shows an example for this step.

\begin{figure}[h!]
  \subcaptionbox{
    A derivation for the string \(\term{scheduled} \term{today}\).
    Light gray arrows show how the bottom-most composition is chained with the monic rules on top.
    \label{fig:dechain:pre}}[\linewidth]{
    \input{figures/derivation-monics-standalone.tex}}

  \vspace{3mm}

  \subcaptionbox{
    The derivation resulting from step\@ \ref{step:short:dechain} applied to the derivation in fig.\@ \ref{fig:dechain:pre}.
    \label{fig:dechain:post}}[\linewidth]{
    \input{figures/derivation-dechained-standalone.tex}}

  \caption{
    Example for step~\ref{step:short:dechain}.
    \label{fig:dechain}}
\end{figure}

\paragraph{Step~\ref{step:short:fuseterm}.}
We replace all occurrences in \(d\) of the form
\begin{itemize}
  \item \(A \to c(B_1, B_2) \Big(B_1 \to (\sigma_1), B_2 \to (\sigma_2)\Big)\) by \(A \to \sem{c}(\sigma_1, \sigma_2)\),
  \item \(A \to c(B_1, B_2) \Big(B_1 \to (\sigma_1), \ldots \Big)\) by \(A \to \sem{c}_1(\sigma_1)\,(B_2)\Big( \ldots \Big)\), and
  \item \(A \to c(B_1, B_2) \Big(\ldots, B_2 \to (\sigma_2) \Big)\) by \(A \to \sem{c}_2(\sigma_2)\,(B_1)\Big( \ldots \Big)\).
\end{itemize}
The removed nonterminals, \(B_1\) and/or \(B_2\), are part-of-speech tags.
We note that in each of the three cases, we construct a lexical rule, and after this step, every rule in $d$ is either branching or lexical.
Moreover, every terminal rule in $d$ is either initial (and uni-lexical) or double-lexical.
Fig.\@ \ref{fig:fuseterm} shows an example for this step.

\begin{figure}[h!]
  \subcaptionbox{
    A derivation for the string tuple \((\term{A} \term{hearing}, \term{on} \term{the} \term{issue})\).
    Light gray arrows show the lexical symbols that are put into binary non-lexical rules during step\@ \ref{step:short:fuseterm}.
    \label{fig:fuseterm:pre}}[\linewidth]{
    \input{figures/derivation-standalone.tex}}

  \vspace{3mm}

  \subcaptionbox{
    The derivation resulting from step\@ \ref{step:short:fuseterm} applied to the derivation in fig.\@ \ref{fig:fuseterm:pre}.
    Light gray arrows show how lexical symbols will be propagated through the derivationt to lexicalize the rule in the root during step\@ \ref{step:short:propterm}.
    \label{fig:fuseterm:post}}[\linewidth]{
    \input{figures/derivation-fused-standalone.tex}}

  \caption{
    Example for step~\ref{step:short:fuseterm}.
    \label{fig:fuseterm}}
\end{figure}

\paragraph{Step~\ref{step:short:propterm}.}
For every occurrence \(r\) of a branching rule $A \to c(A_1, A_2)$ in \(d\), let us consider the occurrence \(t\) of the leftmost terminating rule that is reachable via the second successor of \(r\).
At each node \(s\) on the path from \(t\) to \(r\) (from bottom up):
\begin{itemize}
  \item If \(s\) is \(t\), we remove the leftmost symbol in the rule's composition at \(s\).
  \item
    If \(s\) is neither \(t\) nor \(r\), we insert the last removed symbol right before the variable \(\x_1^1\) and then remove the leftmost symbol in the rule's composition at \(s\).
    If there is a monic rule at node \(s\), we need to store the information if the same symbol was inserted and removed or if the symbol in the rule was swapped.
    In the following, we consider this information as part of the rule, but it has no influence on its nonterminals and composition; it is solely needed for the back transformation described in sec.~\ref{sec:step:propterm}.
  \item If \(s\) is \(r\), we insert the last removed symbol right before the variable \(\x_2^1\) in the rule's composition at \(s\).
\end{itemize}
If, after removal of a symbol, the first component in the composition was empty, we annotate the lhs nonterminal (and the matching rhs nonterminal in the parent) with $^-$ and remove the empty component.
Moreover, let this nonterminal be the $i$th nonterminal in the rhs of the parent;
we also remove $\x_i^1$ in the parent's composition and replace every other occurrence of $\x_i^j$ by $\x_i^{j-1}$.
Otherwise, we annotate the nonterminals with \(^+\).

We note that the rule at $s$ is uni-lexical and branching now, the rule at \(t\) is uni-lexical and terminating, and the number of lexical symbols in each rule between them did not change.
After this step, every rule in $d$ is lexical.
Figs.\@ \ref{fig:fuseterm:post} and \ref{fig:propterm} show an example for this step.

\begin{figure}[h!]
  \centering
  \input{figures/derivation-proped-standalone.tex}
  \caption{
    The derivation resulting from step\@ \ref{step:short:propterm} applied to the derivation in fig.\@ \ref{fig:fuseterm:post}.
    A light gray annotation \emph{swapped} marks a monic rule where the terminal symbol changed.
    \label{fig:propterm}}
\end{figure}
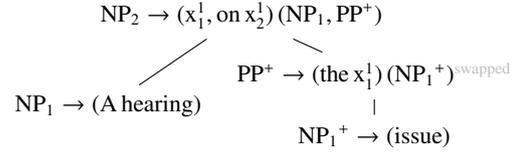

\paragraph{Step~\ref{step:short:split}.}
Each occurrence of the form
  \(A \to (\sigma_1 \sigma_2)\) is replaced by \(A \to (\sigma_1 \x_1) (A^\text{R}) \Big(A^\text{R} \to (\sigma_2)\Big)\), and
  \(A \to (\sigma_1, \sigma_2)\) is replaced by \(A \to (\sigma_1, \x_1) (A^\text{R}) \Big(A^\text{R} \to (\sigma_2)\Big)\),
where \(A^\text{R}\) is a new nonterminal.
After this step, every rule in $d$ is uni-lexical.
Figs.\@ \ref{fig:propterm} and \ref{fig:split} show an example for this step.

\begin{figure}[h!]
  \centering
  \input{figures/derivation-splitted-standalone.tex}
  \caption{
    The derivation resulting from step\@ \ref{step:short:split} applied to the derivation in fig.\@ \ref{fig:propterm}.
    Each rule in the derivation contains exactly one lexical symbol.
    \label{fig:split}}
\end{figure}
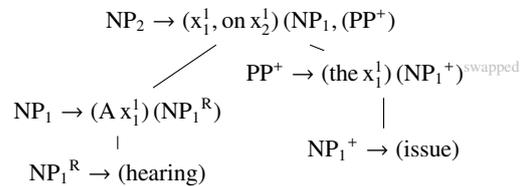

\section{Unlexicalizing Derivations}\label{sec:unlex}

The supertagging-based parser produces derivations of the LCFRS \lexG{} that was introduced in Section~\ref{sec:lex}.
In this section, we describe how they are transformed into equivalent derivations of the LCFRS $G$ that is extracted from the corpus.

Given a derivation $d \in \derivs^{\lexG}$, we have apply to $d$ the inverse transformation of each of the steps~\ref{step:short:split}--\ref{step:short:binarize}.

\paragraph{Inverse of step~\ref{step:short:split}.}
Every subtree of the form $A \to (\sigma_1 \x_1^1)(A^\R) \big(A^\R \to (\sigma_2)\big)$ is replaced by the rule $A \to (\sigma_1 \sigma_2)$.
Similarly, every subtree of the form $A \to (\sigma_1, \x_1^1)(A^\R) \big(A^\R \to (\sigma_2)\big)$ is replaced by the rule $A \to (\sigma_1, \sigma_2)$.

\paragraph{Inverse of step~\ref{step:short:propterm}.}\label{sec:step:propterm}
This step includes a reconstruction of compositions, which is described in detail in app.~\ref{sec:unlex-detail} as it is straight-forward, but includes many case distinctions.
It is applied to each occurrence $r$ of branching rules of the form $A \to c(A_1, A_2)$ from the bottom to the top (i.e.\@ it was already done for branching rules in the subtrees below each node).
Let $t$ be the leftmost occurrence of a terminating rule that is reachable from the second child of $r$.
At each node $s$ on the path from $r$ to $t$ we proceed as follows.
\begin{itemize}
    \item
        If $s$ is $r$, let $\sigma$ be the lexical symbol in $c$.
        We replace $c$ as described in app.~\ref{sec:unlex-detail}.
        Then we pass $\sigma$ to the next node on the path to $t$.
    \item
        If $s$ is neither \(r\) nor \(t\), and there is a branching rule at \(s\), we pass the passed symbol to the next node on the path to \(t\).
        We note that the composition of $s$ was already changed before in this step.
    \item
        If there is a monic rule of the form $B \to c'(B_1)$ at \(s\), we replace \(c'\) as described in app.~\ref{sec:unlex-detail}.
        Let $\sigma_1$ be the passed symbol and $\sigma_2$ the lexical symbol in $c'$.
        If the terminal in this rule was swapped during step~\ref{step:short:propterm}, we pass \(\sigma_2\) to the next node on the path to \(t\), otherwise we pass \(\sigma_1\).
    \item
        If $s$ is $t$, it must be of the form $B \to (\sigma_2)$.
        Let $\sigma_1$ be the lexical symbol received from its parent.
        We replace it by $B \to (\sigma_1, \sigma_2)$ if $B$ is annotated with $^-$ and by $B \to (\sigma_1 \sigma_2)$ otherwise.
\end{itemize}
All annotation is removed from the nonterminals after this step.

\paragraph{Inverse of step~\ref{step:short:fuseterm}.}
We replace every occurrence of a terminating rule of the form
\begin{itemize}
    \item $A \to (\sigma_1 \sigma_2)$ by $A \to (\x_1 \x_2)(A_1, A_2) \big( A_1 \to (\sigma_1), A_2 \to (\sigma_2) \big)$
        and
    \item $A \to (\sigma_1, \sigma_2)$ by $A \to (\x_1, \x_2)(A_1, A_2) \big( A_1 \to (\sigma_1), A_2 \to (\sigma_2) \big)$,
\end{itemize}
where $A_1$ and $A_2$ are the part-of-speech tags of $\sigma_1$ and $\sigma_2$, respectively.

We replace every occurrence of the form $A \to c(B) \big( \dots \big)$, where \(\sigma\) is the lexical symbol in \(c\) and \(A_1\) the part-of-speech tag of \(\sigma\), as follows:
\begin{itemize}
  \item if \(\sigma\) is the first symbol in \(c\), then \(c'\) is obtained from \(c\) by replacing, for each \(j \in [\fanout(B)]\), \(\x_1^j\) with \(\x_2^j\) and \(\sigma\) with \(\x_1^1\);
    the part is replaced by \(A \to c'(A_1, B) \big( A_1 \to (\sigma), \ldots \big)\),
  \item otherwise, \(c'\) is obtained from \(c\) by replacing \(\sigma\) with \(\x_2^1\);
    the part is replaced by \(A \to c'(B, A_1) \big(\ldots, A_1 \to (\sigma) \big)\).
\end{itemize}
The composition \(c'\) is constructed such that \(\sem{c'}_1(\sigma) = c\) in the first case, and \(\sem{c'}_2(\sigma) = c\) in the second case.

We remark that the choice to use part-of-speech tags for the nonterminals $A_1$ and $A_2$ is pragmatic.
It is motivated by the fact that, due to the annotation scheme of the treebank, the nonterminals $B_1$ and $B_2$ removed in step~\ref{step:short:fuseterm} are likely to be the part-of-speech tags of $\sigma_1$ and $\sigma_2$.
This heuristics only fails when the part-of-speech tag is preceded by a chain, e.g.\@ $\nt{NP}\big(\nt{NN}(\term{students})\big)$.
In such a case we would collapse the chain in step~\ref{step:short:dechain} and thus remove the nonterminal $\nt{NP}+\nt{NN}$, losing the information about the chain.
However, occurrences of this kind are so rare that we opted to ignore them.

\paragraph{Inverse of step~\ref{step:short:dechain}.}
We \emph{repeatedly} replace every occurrence of the form \( A\text+B \to c(\dots) \big( \dots \big) \) by
\[
    A \to \id_{\fanout(B)}\,(B)\,\Big( B \to c(\dots) \big( \dots \big) \Big),
\]
until there are no nonterminals of the form $A\text+B$ left in $d$.
If the rule \(A\text+B \to c(\dots)\) has a parent in \(d\), then we replace the nontermial \(A\text+B\) in its rule's rhs by \(A\).

\paragraph{Inverse of steps~\ref{step:short:readoff} and~\ref{step:short:binarize}.}
The derivation is transformed into a (binary) parse tree which is then unbinarized using the standard procedure.

\section{Supertagging}
We perform supertagging-based parsing with uni-lexical LCFRS.
It consists of two phases.

\begin{enumerate*}[label=(\arabic*)]
    \item Given a uni-lexical LCFRS $\lexG$ and sentence, a discriminative model predicts for each sentence position a sample of rules of $\lexG$.
        This phase is called \emph{supertagging} and can be considered as a sequence tagging task.
        The size of the sample for each position is a fixed number $k$ which constitutes a hyperparameter of our approach.
        \\
    \item We construct a new grammar $\lexG'$ from the rules predicted during supertagging.
        We replace the lexical symbol of each rule by the sentence position it was predicted for.
        Then we employ a usual chart-based parsing algorithm to parse the sequence of sentence positions with $\lexG'$.
        As $\lexG'$ has considerably fewer rules than $\lexG$, this approach shifts a huge amount of work from parsing with grammars to predicting the rules.
\end{enumerate*}

\section{Experiments}
\paragraph{Implementation.}
The corpus lexicalization procedure and parsing was implemented as a part of Disco-Dop~\cite{CraSchBod16}, from which we could borrow the LCFRS extraction and parsing implementation.
Moreover, we used the computation of evaluation scores in Disco-Dop.

The sequence tagger for the supertagging algorithm was implemented using the flair framework~\cite{Akb19}.
It features to kinds of word embeddings that we tested:
\begin{itemize}
  \item (bert) the output of the four topmost layers of a pretrained bert model, which is fine-tuned during training,
  \item (bilstm) the concatenation of fasttext~\cite{Mik18} (pretrained language-specific word embeddings), flair~\cite{Akb18} (pretrained language-specific bidirectional subword embeddings, fine-tuned during training) and learned embeddings of gold POS-tags.
\end{itemize}
On top of these embeddings, there are two layers of Bi-LSTMs~\cite{Hoc97} and one linear layer.
The sequence tagger is trained to predict the gold supertag via cross-entropy loss.
More details with respect to hyperparameters for both models are shown in tbl.\@ \ref{tbl:model}.

\begin{table}
  \caption{
    Hyperparameters for the sequence tagger.
    \label{tbl:model}}
  \small
  \begin{center}
    \begin{tabular}{lcc} \toprule
      Parameter                   & Value (bilstm) & Value (bert)                                                      \\ \midrule
      \multirow{2}{*}{embeddings}   & fasttext,               & \multirow{2}{*}{top 4 bert layers}                     \\
                                    & flair                   &                                                        \\
      Bi-LSTM                       & \multicolumn{2}{c}{2 layers, each 512 hidden states}                             \\
      linear layer                  & \multicolumn{2}{c}{no. of supertags}                                             \\
      dropout                       & \multicolumn{2}{c}{\(0.05\)}                                                     \\
      loss                          & \multicolumn{2}{c}{cross entropy}                                                \\
      optimizer                     & \multicolumn{2}{c}{Adam (\(\beta_1 = 0.9, \beta_2=0.999\))}                      \\
      base lr                       & \multicolumn{2}{c}{\(10^{-4}\)}                                                  \\
      \multirow{2}{*}{lr scheduler} & \multicolumn{2}{c}{reduce on plateau}                                            \\
                                    & \multicolumn{2}{c}{(half $\mathrm{lr}$ if dev.\@ loss increases)}                \\
      batch size                    & \multicolumn{2}{c}{32}                                                           \\
      training epochs               & \multicolumn{2}{c}{max.\@ 200, or until \(\mathrm{lr} < 10^{-7}\)}               \\
      \(k\)-best tags               & \multicolumn{2}{c}{10}                                                           \\ \bottomrule
    \end{tabular}
  \end{center}
\end{table}

\paragraph{Data.}
Following \cite{CoaCoh19}, we use three corpora for discontinuous constituent parsing in our evaluations: NeGra~\cite{Skut98}, Tiger~\cite{Brants04}, and a discontinuous version of the Penn treebank~\cite[DPTB]{EvaKal11}.
The corpora were split according to the usual standards into training, development and test sets.\footnote{
  We use the split for NeGra by \citealp{DubKel03}, for Tiger by \cite{Sed13}, and the standard split for DPTB (sections 2--21 for training, 22 for development, 23 for testing).
}
During development, the lexicalization, tagging and parsing were mostly tested and optimized using NeGra.
%
We binarized and markovized (with horizontal context \(\mathrm{h} = 0\), vertical context \(\mathrm{v} = 1\)) each training corpus before extracting the LCFRS and supertags.
We extracted \(3043\) supertags from the training set of NeGra, \(4108\) from Tiger and \(2968\) from DPTB.

\paragraph{Baselines.}
We report labeled F1-scores, obtained from predicted and gold parse trees using Disco-Dop (using the usual parameters in \emph{proper.prm}), for all constituents (\emph{F1}) and all discontinuous constituents (\emph{Dis-F1}).
Additionally to the scores, parse speed is reported in sentences per second (\emph{sent/s}).

Our scores are compared to recent state-of-the-art parsers for discontinuous constituent trees in four categories:
\begin{itemize}
  \item \emph{grammar-based parsers} -- that directly rely on an underlying (probabilistic) LCFRS,
  \item \emph{chart-based parsers} -- that share parsing algorithms with LCFRS, but lack an explicit set of rules,
  \item \emph{transition systems}, and
  \item \emph{neural systems} -- all are other recent parsing approaches.
\end{itemize}

\section{Results}
Tbl.\@ \ref{tbl:k} shows some statistics of our parser on the development sets for different amounts of supertags taken into consideration for each token (\(k\)).
Specifically, we report the parsing speed (\emph{sent/s}), the rate where the gold tag was among the \(k\) predicted tags per token (\emph{tag acc.}) and parsing scores (labeled \emph{precision}, \emph{recall} and \emph{F1}).

We see the parsing speed drops significantly where \(k > 10\), but there are barely any gains in terms of parsing scores.
As expected, the recall increases drastically with rising \(k\).
However, interestingly, the precision drops at first (as expected, but at a much smaller amount), but then also rises slowly.
We found \(k=10\) to be a good parameter for the rest of our experiments.
\begin{table}
  \caption{
    Results for different values for \(k\), i.e.\@ how many supertags for each token are used for parsing, on development sets after training.
    Includes only the results for our model using bert embeddings.
    \label{tbl:k}
  }
  \small\centering
  \begin{tabular}{c|cSSSS}
    \toprule
    \multirow{2}{*}{\(k\)}
          & \multicolumn{5}{c}{NeGra}                 \\
          &{sent/s}&{tag acc.}&{prec.}&{rec.} &{F1}   \\ \midrule
    1     & 44     & 85.08    & 90.76 & 60.89 & 72.88 \\
    2     & 45     & 90.74    & 88.52 & 77.90 & 82.87 \\
    3     & 41     & 92.47    & 88.45 & 83.31 & 85.80 \\
    5     & 39     & 94.04    & 88.48 & 86.74 & 87.60 \\\rowcolor{lightlightgray}
    10    & 34     & 95.76    & 88.63 & 88.31 & 88.47 \\
    15    & 11     & 96.46    & 88.76 & 88.56 & 88.66 \\
    20    & 1      & 97.03    & 88.64 & 88.53 & 88.64 \\ \midrule
    \multirow{2}{*}{\(k\)}
          & \multicolumn{5}{c}{Tiger}                 \\
          &{sent/s}&{tag acc.}&{prec.}&{rec.} &{F1}   \\ \midrule
    1     & 42     & 88.66    & 93.07 & 74.20 & 82.57 \\
    2     & 41     & 93.44    & 91.74 & 85.96 & 88.75 \\
    3     & 40     & 94.86    & 91.35 & 88.54 & 89.92 \\
    5     & 40     & 96.15    & 91.46 & 90.18 & 90.82 \\\rowcolor{lightlightgray}
    10    & 32     & 97.47    & 91.52 & 90.92 & 91.22 \\
    15    & 7      & 98.04    & 91.58 & 91.03 & 91.30 \\
    20    & {--}   & {--}     & {--}  & {--}  & {--}  \\ \midrule
    \multirow{2}{*}{\(k\)}
          & \multicolumn{5}{c}{DPTB}                  \\
          &{sent/s}&{tag acc.}&{prec.}&{rec.} &{F1}   \\ \midrule
    1     & 39     & 90.32    & 92.61 & 69.40 & 79.34 \\
    2     & 42     & 94.20    & 91.36 & 83.95 & 87.50 \\
    3     & 35     & 95.38    & 91.10 & 87.19 & 89.11 \\
    5     & 32     & 96.47    & 91.50 & 89.84 & 90.67 \\\rowcolor{lightlightgray}
    10    & 25     & 97.55    & 91.48 & 91.33 & 91.41 \\
    15    & 6      & 98.06    & 91.53 & 91.51 & 91.52  \\
    20    & {--}   & {--}     & {--}  & {--}  & {--}  \\ \bottomrule
  \end{tabular}
\end{table}

Publications for supertaggers often omit parsing scores and only give accuracies of their predictions with respect to gold supertags.
Tbl.\@ \ref{tbl:accuracy} compares our supertagger to some recent publications.
We include some results for CCGBank that we cannot compare to, but it shows how well supertagging performs for combinatory categorial grammars (CCG).

\begin{table*}
  \caption{
    Our results compared to other published supertaggers.
    \Citealp{Bla18} used a slightly different split of Tiger.
    \label{tbl:accuracy}
  }
  \centering\small
  \begin{tabular}{ll|cS|cS|cS|cS}
  \toprule
  \multirow{2}{*}{Model} & \multirow{2}{*}{formalism}
                                    & \multicolumn{2}{c|}{NeGra}
                                                       & \multicolumn{2}{c|}{Tiger}
                                                                          & \multicolumn{2}{c|}{DPTB}
                                                                                              & \multicolumn{2}{c}{CCGBank} \\
     &                              & tags & accuracy  & tags & accuracy  & tags & accuracy   & tags & accuracy       \\\midrule
  \citealp{Bla18}        & TAG      & {--} & {--}      & 3426 & 88.51     & {--} & {--}       & {--} & {--}       \\
  \citealp{Kad18}        & CCG      & {--} & {--}      & {--} & {--}      & {--} & {--}       & 1284 & 94.49      \\
  \citealp{Kas17}        & TAG      & {--} & {--}      & {--} & {--}      & 4727 & 89.71      & {--} & {--}       \\\rowcolor{lightlightgray}
  ours (bilstm)          & LCFRS    & 3275 & 79.98     & 4614 & 78.80     & 4509 & 84.76      & {--} & {--}       \\\rowcolor{lightlightgray}
  ours (bert)            & LCFRS    & 3275 & 86.93     & 4614 & 84.28     & 4509 & 90.78      & {--} & {--}       \\
  \citealp{vaswani2016supertagging}
                         & CCG      & {--} & {--}      & {--} & {--}      & {--} & {--}       & 1284 & 94.5       \\ \bottomrule
  \end{tabular}
\end{table*}

Tbl.\@ \ref{tbl:results} shows the parsing scores and speed of our final model on the test set compared to the scores reported in other recent publications for discontinuous constituent parsing.
The experiments suggest that parsing using LCFRS can greatly benefit from supertagging, with respect to, both, speed and accuracy.
This, however requires a strong discriminative classifier for the sequence tagger to predict useful rules.

Compared to other parsing approaches, we obtain results that on par with state-of-the-art parsing approaches, we would like to especially highlight the excellent scores for discontinuous constituents.
This suggests that grammar-based approaches can keep up with other parsers, if they are appropriately combined with discriminative models.
Recent publications of \citealp{Cor20} and \cite{StaSte20} obtain similar results using approaches that base on parsing with LCFRS, but lack of an underlying grammar and solely rely on neural network predictions.

\begin{table*}
  \defcitealias{FerGom20a}{Fern{\'a}ndez-G. and G{\'o}mez-R., 2020}
  \defcitealias{VilGom20}{Vilares and Gómez-R., 2020}
  \caption{
    Our results on test sets compared to other published constituent parsers.
    \Citealp{CraSchBod16} use another split for the Tiger corpus.
    \label{tbl:results}
  }
  \small
  \begin{tabular}{l|SSc|SSc|SSc}
    \toprule
    \multirow{2}{*}{Model}           & \multicolumn{3}{c|}{NeGra}
                                                               & \multicolumn{3}{c|}{Tiger}
                                                                                         & \multicolumn{3}{c}{DPTB} \\
                                     &  {F1} &{Dis-F1}&{sent/s}&  {F1} &{Dis-F1}&{sent/s}&  {F1} &{Dis-F1}&{sent/s} \\
    \midrule                         \multicolumn{10}{c}{Grammar-based systems}                                    \\\midrule
    \citealp{CraSchBod16}            & 76.8  &   {--} &      2 & 78.2  &   {--} &      1 & 87.0  &   {--} &   $<1$ \\
    \citealp{Geb20}                  & 81.7  &  43.5  &   {--} & 77.7  &  40.7  &   {--} &  {--} &   {--} &   {--} \\\rowcolor{lightlightgray}
    ours (bilstm)                    & 82.79 &  52.93 &     53 & 81.82 &  54.58 &     39 & 89.07 &  63.36 &     47 \\\rowcolor{lightlightgray}
    ours (bert)                      & 89.01 &  70.09 &     30 & 86.81 &  66.00 &     24 & 92.87 &  72.73 &     36 \\
    \citealp{Ver16}                  & {--}  &{--}    &   {--} & 79.50 &   {--} &   {--} &  {--} &   {--} &   {--} \\
    \midrule                         \multicolumn{10}{c}{Chart-based systems}                                      \\\midrule
    \citealp{Cor20} (w/o bert)       & 86.3  &  56.1  &   {--} & 85.2  &  51.2  &   {--} & 92.9  &  64.9  &   {--} \\
    \citealp{Cor20} (w/  bert)       & 91.6  &  66.1  &   {--} & 90.0  &  62.1  &   {--} & 94.8  &  68.9  &   {--} \\
    \citealp{StaSte20}               & 83.3  &  50.7  &   {--} & 83.4  &  53.5  &   {--} & 90.5  &  67.7  &   {--} \\
    \midrule                         \multicolumn{10}{c}{Transition systems}                                       \\\midrule
    \citealp{CoaCoh19}               & 84.0  &  54.0  &   {--} &  87.6  &  52.5 &   {--} & 91.4  &  70.9  &   {--} \\
    \citealp{CoaCraCoh19}            & 83.2  &  54.6  &   {--} &  82.7  &  55.9 &    126 & 91.0  &  71.3  &    80  \\
    \midrule                         \multicolumn{10}{c}{Neural systems}                                           \\\midrule
    \citetalias{FerGom20a}           & 86.1  &  59.9  &   {--} &  86.3  &  60.7 &   {--} &  {--} &   {--} &   {--} \\
    \citetalias{VilGom20} (bilstm)   & 77.1  &  36.5  &    715 &  79.2  &  40.1 &    568 & 89.1  &  41.8  &    611 \\
    \citetalias{VilGom20} (bert)     & 84.2  &  46.9  &     81 &  84.7  &  51.6 &     80 & 91.7  &  49.1  &     80 \\
    \bottomrule
  \end{tabular}
\end{table*}

\section{Conclusion}
We described an approach to utilize supertagging for parsing discontinuous constituent trees with LCFRS and implemented it.
Compared to other parsers for the same grammar formalism, we achieve state of the art results, i.e.\@ we are more accurate and also faster (cf.\@ tbl.\ \ref{tbl:results}, Grammar-based systems).
In contrast to previous parsers utilizing LCFRS, we can even keep up with other recent parsing approaches and achieve excellent results for discontinuous constituents (cf.\@ tbl.\ \ref{tbl:results}, columns for Dis-F1).

\paragraph{Future Work.}
Disco-Dop currently only supports parsing sentences up to 128 words and, unfortunately, a bug was keeping us from increasing this limit.
Resolving this issue should increase the scores slightly in Tiger and DPTB.

The sequence tagger's hyperparameters need to be addressed in a proper parameter search.
Now, they are mostly set by trial and error.

\section*{Acknowledgements}
We thank Alex Ivliev for conducting early experiments during the development of our parser.

\bibliography{references}
\bibliographystyle{acl_natbib}

\appendix

\section{Supplementary details on constructions}
\subsection{Unlexicalizing Derivations}%
\label{sec:unlex-detail}

In this appendix we describe how the original compositions of branching and monic rules are computed when applying the inverse of step~\ref{step:short:propterm} to some derivation $d$.

\paragraph{Branching rules.}
Let $r$ be a branching rule of the form $A \to (u_1, \ldots, u_s)\,(A_1, A_2)$ and $\sigma$ be the lexical symbol in $(u_1, \ldots, u_s)$.
\begin{itemize}
  \item If $A_2$ is annotated with $^-$ (i.e., its first component was removed during step~\ref{step:short:propterm}), we replace $\sigma$ with  $\x_2^0$, and replace every occurrence of $\x_2^i$ by $\x_2^{i+1}$.
  \item Otherwise, \(\sigma\) is removed from \((u_1, \ldots, u_s)\).
\end{itemize}

Moreover, if $r$ occurs as a successor of the right child of some other branching rule in $d$, then the nonterminals $A$ and $A_1$ are annotated as well.
\begin{itemize}
  \item If \(A_1\) and \(A\) are annotated with \(^-\), then we replace \((u_1, \ldots, u_s)\) by \((\x_1^0, u_1, \ldots, u_s)\).
  \item If \(A_1\) is annotated with \(^-\) and \(A\) with \(^+\), then we replace \((u_1, \ldots, u_s)\) by \((\x_1^0 u_1, \ldots, u_s)\).
\end{itemize}
After that, we replace every occurrence of $\x_1^i$ by $\x_1^{i+1}$.

We recall that during step~\ref{step:short:propterm}, every occurrence of a branching rule is processed up to two times:
once for inserting a lexical symbol and potentially a second time for propagating a lexical symbol to its parent.
Since both modifications affect the variables of different components, they can be undone independently from each other (as described in Section~\ref{sec:unlex}) or at the same time (as done here).

\paragraph{Monic rules.}
Let $A \to c(A_1)$ be a monic rule, $\sigma_1$ be the lexical symbol received from its parent, and $\sigma_2$ be the lexical symbol of $c$. 

\begin{enumerate*}[label=(\arabic*)]
    \item If the terminal of this rule was swapped during step~\ref{step:short:propterm} and $A_1$ is annotated with $^-$, we insert $\x_1^0$ before $\x_1^1$ in $c$ and replace every occurrence of $\x_1^i$ by $\x_1^{i+1}$.
        Moreover, if $A$ is annotated with $^-$, a component split is added after $\x_1^1$.
        \\
    \item Otherwise, we replace $\sigma_2$ in $c$ by $\sigma_1$.
        If $A_1$ is annotated with $^-$, we insert $\x_1^0$ after $\sigma_1$ in $c$ and replace every occurrence of $\x_1^i$ by $\x_1^{i+1}$.
        Moreover, if $A$ is annotated with $^-$, a component split is added after $\sigma_1$.
\end{enumerate*}

\end{document}

%% file: preambles/base.tex
\usepackage{adjustbox}
\usepackage{algorithm}
\usepackage[noend]{algpseudocode}
\usepackage{amsmath}
\usepackage{amssymb}
\usepackage{amsthm}
\usepackage[english]{babel}
\usepackage{csquotes}
\usepackage{ellipsis}
\usepackage{etaremune}
\usepackage{isomath}
\usepackage[inline]{enumitem}
\usepackage{mathtools}
\usepackage{microtype}
\usepackage{stmaryrd}
\usepackage[
    backgroundcolor=orange!5,
    bordercolor=orange,
    linecolor=orange,
    textsize=footnotesize,
]{todonotes}

\DeclareMathSymbol{:}{\mathpunct}{operators}{"3A}


%% file: preambles/mathops.tex
\newcommand{\DN}{\mathbb{N}}

\DeclareMathOperator{\id}{id}
\DeclareMathOperator{\sort}{sort}
\DeclareMathOperator{\rank}{rk}
\DeclarePairedDelimiter\sem{\llbracket}{\rrbracket}

\DeclareMathOperator{\sTrees}{T}

\newcommand{\x}{\mathrm{x}}
\newcommand{\C}{\mathrm{C}}

\DeclareMathOperator{\fanout}{fo}

\DeclareMathOperator{\derivs}{D}

\newcommand\lexG{\ensuremath{G_{\mathrm{lex}}}}

\newcommand\R{\mathrm R}


%% file: preambles/theorems.tex
\usepackage{thmtools}

\theoremstyle{definition}

\declaretheorem[numbered=no,name=Lemma]{lemma*}

%% file: preambles/tikz.tex
\usepackage{tikz}

\usetikzlibrary{calc}
\usetikzlibrary{fit}
\usetikzlibrary{graphs}
\usetikzlibrary{positioning}
\usetikzlibrary{tikzmark}

\newcommand*\nt[1]{\ensuremath{\mathrm{#1}}}
\newcommand*\term[1]{\ensuremath{\operatorname{#1}}}

\tikzset{>=stealth}

\makeatletter
\@ifclassloaded{beamer}{%
\newcounter{jjumping}
\resetcounteronoverlays{jjumping}
\tikzset{
  stop jumping/.style={
    execute at end picture={%
      \stepcounter{jjumping}%
      \immediate\write\pgfutil@auxout{%
        \noexpand\jump@setbb{\the\value{jjumping}}{\noexpand\pgfpoint{\the\pgf@picminx}{\the\pgf@picminy}}{\noexpand\pgfpoint{\the\pgf@picmaxx}{\the\pgf@picmaxy}}
      },
      \csname jump@\the\value{jjumping}@maxbb\endcsname
      \path (\the\pgf@x,\the\pgf@y);
      \csname jump@\the\value{jjumping}@minbb\endcsname
      \path (\the\pgf@x,\the\pgf@y);
    },
  }
}
\def\jump@setbb#1#2#3{%
  \@ifundefined{jump@#1@maxbb}{%
    \expandafter\gdef\csname jump@#1@maxbb\endcsname{#3}%
  }{%
    \csname jump@#1@maxbb\endcsname
    \pgf@xa=\pgf@x
    \pgf@ya=\pgf@y
    #3
    \pgfmathsetlength\pgf@x{max(\pgf@x,\pgf@xa)}%
    \pgfmathsetlength\pgf@y{max(\pgf@y,\pgf@ya)}%
    \expandafter\xdef\csname jump@#1@maxbb\endcsname{\noexpand\pgfpoint{\the\pgf@x}{\the\pgf@y}}%
  }
  \@ifundefined{jump@#1@minbb}{%
    \expandafter\gdef\csname jump@#1@minbb\endcsname{#2}%
  }{%
    \csname jump@#1@minbb\endcsname
    \pgf@xa=\pgf@x
    \pgf@ya=\pgf@y
    #2
    \pgfmathsetlength\pgf@x{min(\pgf@x,\pgf@xa)}%
    \pgfmathsetlength\pgf@y{min(\pgf@y,\pgf@ya)}%
    \expandafter\xdef\csname jump@#1@minbb\endcsname{\noexpand\pgfpoint{\the\pgf@x}{\the\pgf@y}}%
  }
}
}{}
\makeatother

%% file: figures/discontinuous.tex
\begin{figure*}
    \renewcommand*\term[1]{\strut\ensuremath{\operatorname{#1}}}%
    \centering
    \begin{tikzpicture}[edge from parent path={(\tikzparentnode.south) -- ++(0,-1.25mm) -| (\tikzchildnode.north)},level distance=7mm,font=\small,sibling distance=10mm,baseline=(c)]
        \coordinate (c) at (0,-39mm);
        \node {\nt{VP}}
        child {
            node[xshift=-1cm,yshift=-28mm] {\nt{VBZ}}
            child {
                node {\term{is}}
            }
        }
        child {
            node {\nt{VP}}
            child {
                node[yshift=-21mm] {\nt{VBN}}
                child {
                    node {\term{scheduled}}
                }
            }
            child {
                node {\nt{NP}}
                child {
                    node[xshift=-3cm] {\nt{NP}}
                    child {
                        node[yshift=-7mm] {\nt{DT}}
                        child {
                            node {\term{A}}
                        }
                    }
                    child {
                        node[yshift=-7mm] {\nt{NN}}
                        child {
                            node {\term{hearing}}
                        }
                    }
                }
                child {
                    node {\nt{PP}}
                    child {
                        node[yshift=-7mm] {\nt{IN}}
                        child {
                            node {\term{on}}
                        }
                    }
                    child {
                        node {\nt{NP}}
                        child {
                            node {\nt{DT}}
                            child {
                                node {\term{the}}
                            }
                        }
                        child {
                            node {\nt{NN}}
                            child {
                                node {\term{issue}}
                            }
                        }
                    }
                }
            }
            child {
                node[xshift=1.5cm] {\nt{NP}}
                child {
                    node[yshift=-14mm] {\nt{NN}}
                    child {
                        node {\term{today}}
                    }
                }
            }
        };
    \end{tikzpicture}
    \hspace{0cm plus 1fil minus 1cm}%
    \renewcommand*\term[1]{\ensuremath{\operatorname{#1}}}%
    \begin{tikzpicture}[font=\small,level distance=10mm]
        \node {$\nt{NP_2} \to (\x_1^1, \x_2^1)\,(\nt{NP_1}, \nt{PP})$}
        child[sibling distance=38mm] {
            node {$\nt{NP_1} \to (\x_1^1 \x_2^1)\,(\nt{DT}, \nt{NN})$}
            child[sibling distance=19mm] {
                node {\strut$\nt{DT} \to (\term{A})$}
            }
            child[sibling distance=19mm] {
                node {$\nt{NN} \to (\term{hearing})$}
            }
        }
        child[sibling distance=38mm] {
            node {$\nt{PP} \to (\x_1^1 \x_2^1)\,(\nt{IN}, \nt{NP_1})$}
            child[sibling distance=19mm] {
                node {\strut$\nt{IN} \to (\term{on})$}
            }
            child[sibling distance=29mm] {
                node {\strut$\nt{NP_1} \to (\x_1^1 \x_1^2)\,(\nt{DT}, \nt{NN})$}
                child[sibling distance=30mm] {
                    node {$\nt{DT} \to (\term{the})$}
                }
                child[sibling distance=10mm] {
                    node {$\nt{NN} \to (\term{issue})$}
                }
            }
        };
    \end{tikzpicture}
    \caption{A discontinuous phrase structure tree of the sentence \emph{A hearing is scheduled on the issue today} (left) and a corresponding LCFRS derivation of the discontinuous noun phrase \emph{A hearing on the issue} (right).\label{fig:discontinuous}}
\end{figure*}

%% file: figures/derivation-monics-standalone.tex
  \begin{tikzpicture}[font=\footnotesize,level 1/.style={sibling distance=30mm},level distance=8mm, remember picture]
    \node {$\nt{VP_2}\texttt{|<>} \to (\x_1^1 \x_2^1) (\nt{VBN}, \nt{NP_1})$}
    child {
      node {$\nt{VBN} \to (\term{scheduled})$}
    }
    child[sibling distance=35mm] {
      node {$\nt{NP_1} \subnode{nt2dest}{\strut} \to \subnode{nplocvar2}{(\x_1^1)} (\nt{NP\text-LOC})$}
      child {
        node {$\subnode{nt2}{\strut \nt{NP\text-LOC}} \subnode{nt1dest}{\strut} \to \subnode{nplocvar}{(\x_1^1)}(\nt{NN})$}
        child {
          node {$\subnode{nt1}{\strut \nt{NN}} \to \subnode{nplocterm}{(\term{today})}$}
        }
      }
    };
    
    \begin{scope}[->, dashed, lightgray]
      \draw (nplocterm) -- (nplocvar.base);
      \draw (nplocvar.north) .. controls ++(-0.3,0.3) and ++(0,-0.3) .. (nplocvar2.base);
      \draw (nt1.north) .. controls ++(0.1,0.3) and ++(0,-0.3) .. node[left] {\(+\)} (nt1dest.base);
      \draw (nt2.north) .. controls ++(0,0.3) and ++(0,-0.3) .. node[left] {\(+\)} (nt2dest.base);
    \end{scope}
  \end{tikzpicture}

%% file: figures/derivation-dechained-standalone.tex
  \begin{tikzpicture}[font=\footnotesize,level 1/.style={sibling distance=20mm}, remember picture]
    \node {$\nt{VP_2}\texttt{|<>} \to (\x_1^1 \x_2^1) (\nt{VBN}, \nt{NP_1\text+NP\text-LOC\text+NN})$}
      child[level distance=8mm] { node {$\nt{VBN} \to (\term{scheduled})$} }
      child[level distance=14mm] { node {$\nt{NP_1\text+NP\text-LOC\text+NN} \to (\term{today})$} };
  \end{tikzpicture}

%% file: figures/derivation-standalone.tex
    \begin{tikzpicture}[font=\small,level distance=8mm, remember picture, every subnode/.style={inner sep=0pt}]
      \coordinate (c) at (0,-17.5mm);
      \node {$\nt{NP} \to (\x_1^1, \x_2^1)\,(\nt{NP}, \nt{PP})$}
        [sibling distance=28mm]
        child[level distance=11mm] {
          node {$\nt{NP} \to (\subnode{d1x11}{\x_1^1} \subnode{d1x21}{\x_2^1})\,(\nt{DT}, \nt{NN})$}
          [level distance=12mm, sibling distance=18mm]
          child { node {$\nt{DT} \to (\subnode{d11t}{\term{A}})$} }
          child { node {$\nt{NN} \to (\subnode{d12t}{\term{hearing}})$} }
        }
        child[level distance=7mm] {
          node {$\nt{PP} \to (\subnode{d2x11}{\x_1^1} \x_2^1)\,(\nt{IN}, \nt{NP_1})$}
          [sibling distance=24mm]
          child[level distance=8mm] { node {\strut$\nt{IN} \to \subnode{d21t}{(\term{on})}$} }
          child[level distance=8mm] { node {\strut$\nt{NP_1} \to (\subnode{d22x11}{\x_1^1} \subnode{d22x21}{\x_2^1})\,(\nt{DT}, \nt{NN})$}
            [sibling distance=18mm]
            child { node {\strut$\nt{DT} \to \subnode{d221t}{(\term{the})}$} }
            child { node {$\nt{NN} \to \subnode{d222t}{(\term{issue})}$} }
          }
        };
      
      \begin{scope}[lightgray, dashed, ->, font=\scriptsize]
        \draw
        (d21t.north)  .. controls ++(0,0.2) and ++(0,-0.6) .. 
          (d2x11.south);
        \draw
        (d221t.north) .. controls ++(.3,0.3) and ++(0,-0.4) .. 
          (d22x11.south);
        \draw
        (d222t.north) .. controls ++(0,0.3) and ++(0,-0.7) .. 
          (d22x21.south);
        \draw
          (d11t.north) .. controls ++(0,0.3) and ++(0,-0.7) .. 
          (d1x11.south);
        \draw
          (d12t.north) .. controls ++(0,0.3) and ++(0,-0.7) .. 
          (d1x21.south);
      \end{scope}
    \end{tikzpicture}

%% file: figures/derivation-fused-standalone.tex
  \begin{tikzpicture}[font=\small,level 1/.style={sibling distance=30mm},level distance=10mm,remember picture,every subnode/.style={inner sep=0pt}]
    \node {$\nt{NP_2} \to (\subnode{dex11}{\vphantom{t}}\x_1^1, \subnode{dex21}{}\x_2^1)\,(\nt{NP_1}, \nt{PP})$}
    child {
      node {$\nt{NP_1} \to (\subnode{de1t}{\term{A}} \term{hearing})$}
    }
    child {
      node {$\nt{PP} \to (\subnode{de2t}{\strut\term{on}}\, \subnode{de2x11}{}\x_1^1)\,(\nt{NP_1})$}
      child {
        node {$\nt{NP_1} \to (\subnode{de21t}{\term{the}} \term{issue})$}
      }
    };
    
    \begin{scope}[->, lightgray, dashed]
      \draw
      (de2t)
      .. 
      controls ++(0,.5) and ++(0,-0.5)
      .. 
      (dex21.center);
      \draw
      (de21t)
      .. 
      controls ++(0,0.5) and ++(0,-0.5)
      .. 
      (de2x11.center);
    \end{scope}
  \end{tikzpicture}

%% file: figures/derivation-proped-standalone.tex
  \begin{tikzpicture}[font=\small,level 1/.style={sibling distance=35mm},level distance=8mm, remember picture]
    \node {$\nt{NP_2} \to (\x_1^1, \term{on} \x_2^1)\,(\nt{NP_1}, \nt{PP}^+)$}
    child[level distance=12mm] {
      node {$\nt{NP_1} \to (\term{A} \subnode{d1t}{\term{hearing}})$}
    }
    child {
      node {$\nt{PP}^+ \to (\term{the} \x_1^1)\,(\nt{NP_1}^+)^{\text{\color{lightgray} swapped}}$}
      child {
        node {$\nt{NP_1}^+ \to (\term{issue})$}
      }
    };
  \end{tikzpicture}

%% file: figures/derivation-splitted-standalone.tex
  \begin{tikzpicture}[font=\small,level 1/.style={sibling distance=35mm},level distance=7mm, remember picture]
    \node {$\nt{NP_2} \to (\x_1^1, \term{on} \x_2^1)\,(\nt{NP_1}, (\nt{PP}^+)$}
    child[level distance=12mm] {
      node {$\nt{NP_1} \to (\term{A} \x_1^1)\,(\nt{NP_1}^\text{R})$}
      child[level distance=8mm] { node {$\nt{NP_1}^\text{R} \to (\term{hearing})$} }
    }
    child {
      node {$\nt{PP}^+ \to (\term{the} \x_1^1)\,(\nt{NP_1}^+)^{\text{\color{lightgray} swapped}}$}
      child[level distance=10mm] {
        node {$\nt{NP_1}^+ \to (\term{issue})$}
      }
    };
  \end{tikzpicture}